\documentclass[letterpaper, 10 pt, conference]{ieeeconf}  

\usepackage{graphicx}
\usepackage{algorithm}
\usepackage{algorithmic}
\usepackage{caption}
\usepackage{subcaption}
\usepackage{mathtools}
\usepackage{mathrsfs, amsmath} 
\usepackage{color}
\usepackage{array}
\newcolumntype{P}[1]{>{\centering\arraybackslash}p{#1}}
\newcolumntype{M}[1]{>{\centering\arraybackslash}m{#1}}
\title{
    Active Adversarial Evader Tracking with a Probabilistic Pursuer \\ under the Pursuit-Evasion Game Framework
}

\author{
\authorblockN{Varun Chandra Jammula}
\authorblockA{
vjammula@asu.edu}
\and
\authorblockN{Anshul Rai}
\authorblockA{
arai10@asu.edu}
\and
\authorblockN{Yezhou Yang}
\authorblockA{
yz.yang@asu.edu}
}

\begin{document}

\maketitle

\begin{abstract}

Given a mapped environment, we formulate the problem of visually tracking and following an evader using a probabilistic framework. In this work, we consider a non-holonomic robot with a limited visibility depth sensor in an indoor environment with obstacles. The mobile robot that follows the target is considered a pursuer and the agent being followed is considered an evader. We propose a probabilistic framework for both the pursuer and evader to achieve their conflicting goals. We introduce a smart evader that has information about the location of the pursuer. The goal of this variant of the evader is to avoid being tracked by the pursuer by using the visibility region information obtained from the pursuer, to further challenge the proposed smart pursuer. 
To validate the efficiency of the framework, we conduct several experiments in simulation by using Gazebo and evaluate the success rate of tracking an evader in various environments with different pursuer to evader speed ratios.
Through our experiments we validate our hypothesis that a smart pursuer tracks an evader more effectively than a pursuer that just navigates in the environment randomly. We also validate that an evader that is aware of the actions of the pursuer is more successful at avoiding getting tracked by a smart pursuer than a random evader.
Finally, we empirically show that while a smart pursuer does increase it's average success rate of tracking compared to a random pursuer, there is an increased variance in its success rate distribution when the evader becomes aware of its actions.

\end{abstract}

\section{Introduction}

Given a mapped environment and an adversarial evader traversing the environment, we develop a probabilistic model for an autonomous non-holonomic mobile robot that maximizes the probability of tracking the evader in the environment, dubbed as a smart pursuer. It is an interesting yet challenging problem because it incorporates several aspects of active perception such as object detection \cite{redmon_you_2015}, object tracking \cite{wu_online_2013}, and planning for autonomous navigation by avoiding obstacles \cite{topiwala_frontier_2018} in addition to our primary task of keeping the evader in the line of sight. The direct applications of our work can be autonomous discreet policing/patrolling and intelligent adversary shadowing, to improve the level of security from both public and private perspectives. Other potential applications range from assisting with carrying luggage to intelligent environmental monitoring.


In this work, we consider a pursuer robot equipped with a limited visibility depth sensor whose goal is to actively track an evader that is moving in an environment with obstacles. The pursuer has prior knowledge of the workspace in the form of an occupancy grid map obtained through SLAM \cite{durrant-whyte_simultaneous_2006}. The goal of the evader is to avoid being tracked by the pursuing robot while traversing in the environment. Essentially, the pursuer robot and evader form the two competing opponents in a pursuit-evasion game.



Current research methods focus on various aspects of the pursuit-evasion game. Some methods focus on decomposing the environment into various forms such as an information graph\cite{gerkey_visibility-based_2006} using visibility of the pursuer and voronoi tessellation of the environment \cite{pierson_intercepting_2017} to reduce the required actions taken by the pursuer. This decomposition in most cases requires offline processing, and the overall complexity is dependent on the size of the map. Moreover, the majority of related methods proposed in pursuit-evasion formulate the problem as capturing an evader, but we model it as a tracking challenge. Capturing an evader does not necessarily guarantee that the evader was tracked for the maximal amount of time (higher tracking success rate). Further more, tracking an evader is additionally challenging when the evader can access every location in an environment without any restrictions, and the evader is able to access the pursuer's location and visibility region information in an adversarial manner (smart evader). In our experimental setting, we create an indoor environment with a lot of fixed obstacles to hinder the maneuverability for the agents. 


\begin{figure}[t!]
    \centering
    \includegraphics[width=0.45\textwidth]{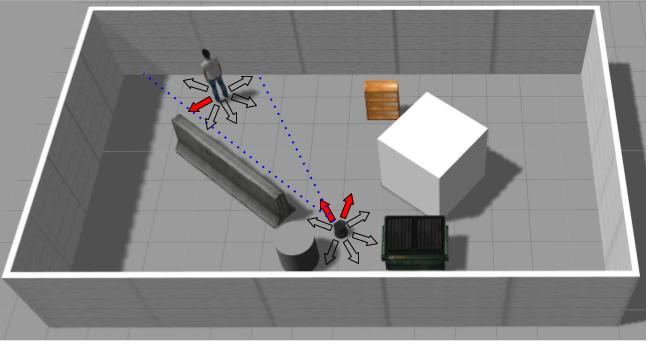}
    \caption{A scenario of the Pursuit-Evasion game where the human is the evader and TurtleBot is the pursuer. The blue lines represent the boundaries of the line-of-sight of the pursuer. The transparent arrows represent the possible random actions the agents can take. The red arrows represent the smart actions an agent can take to fulfill its goal of evading or tracking the opponent.}
    \label{fig:1}
\end{figure}

Given the map of the environment as an occupancy grid map \cite{thrun_robotic_2003}, and assuming the evader is in the line of sight of the pursuer at the beginning, we model our problem as a probabilistic tracking game where pursuer and evader have opposing roles (as shown in Fig.~\ref{fig:1}). Both the pursuer and evader are bound by their velocities $v_p$ and $v_e$. When the pursuer detects the evader, it employs particle filter based visual tracking \cite{bouguet_pyramidal_2000} to track the evader in the current view and uses the depth information from the depth sensor to estimate the position of the evader. When the pursuer cannot track or detect the evader in the image plane, it once again employs a particle filtering based localization module to estimate the state of the evader on the grid map. After estimating the position of the evader in the map, the pursuer moves to the location yielded by the weighted sum of particles, which leads to a higher chance of tracking the evader. 
We present a smart pursuer that maximizes the probability of maintaining the evader in its field of view and probabilistic evader that minimizes the probability of being tracked by the pursuer.


To evaluate the efficacy and the efficiency of our smart pursuer model, we evaluate the success rate of tracking an evader in various environments from various random start points. We also introduce a smart evader whose goal is to actively avoid being tracked by the pursuer. The smart evader knows the location of the pursuer at all time and by using the visibility region of the pursuer, it computes a goal location that minimizes the probability of it being tracked. 
In order to test how reliable and robust our pursuer model is, we investigate the effect of speed ratios and also test the complexity of environments in which the robot will navigate. Moreover, to validate that our pursuer model can capture smart evaders we investigate the effect of various types of evader behavior based on the tracking success rate. Our experimental results observed validate that: 1) Smart pursuer performs better at tracking an evader than a random pursuer; 2) The average success rate of tracking a smart evader is less than the average success rate of tracking a random evader; 3) In the case of having a smart evader, our smart pursuer is able to track the evader with a reasonably high success rate, and in the majority of the scenarios observed, our pursuer has over $45\%$ success rate.   Our contributions to the field in this work thus can be summarized as follows:
\begin{enumerate}
  \item We propose a new pursuit-evasion game setting which aims at better tracking rates, and we show that this setting has many applications.
   \item We set up a virtual environment, with controllable hyper-parameters (such as pursuer/evader speed ratio, environment settings, and evader behavior models) to formally validate our proposed system. We intend to make the testing bed public as well as the testing protocols available upon the publication of this draft for the community to conduct further research.
   \item We propose a probabilistic framework to optimize the chance of the pursuer tracking both the random and smart evaders, and we validate its efficacy and efficiency on both simulation and physical platforms.
\end{enumerate}



\section{Methodology}

\subsection{Problem Statement}

We consider the problem of tracking a mobile evader with a single mobile pursuer. The game is played in discrete intervals of 1 unit each for a limited amount of time $t_{max}$ and hence $\tau = \{0, 1, 2 .. ,k, ..t_{max}\}$.
Let $x^p_k, x^e_k$ be the states and $z^p_k, z^e_k$ be the observations of pursuer and evader respectively at time step $k$. Each $x_k$ is a triplet of the agent's coordinates on the map and also the orientation $x_k = (x, y, \theta)$.

The pursuer agent we consider is non-holonomic with a limited Field Of View (FOV) (in our experiments, we set it to $\pi/3$) and a fixed, front-facing camera. We analyze how the tracking performance of the pursuer is affected by the independent behavior of the evader and pursuer agents.

\begin{figure*}[h]
    \centering
    \includegraphics[width=6.5in]{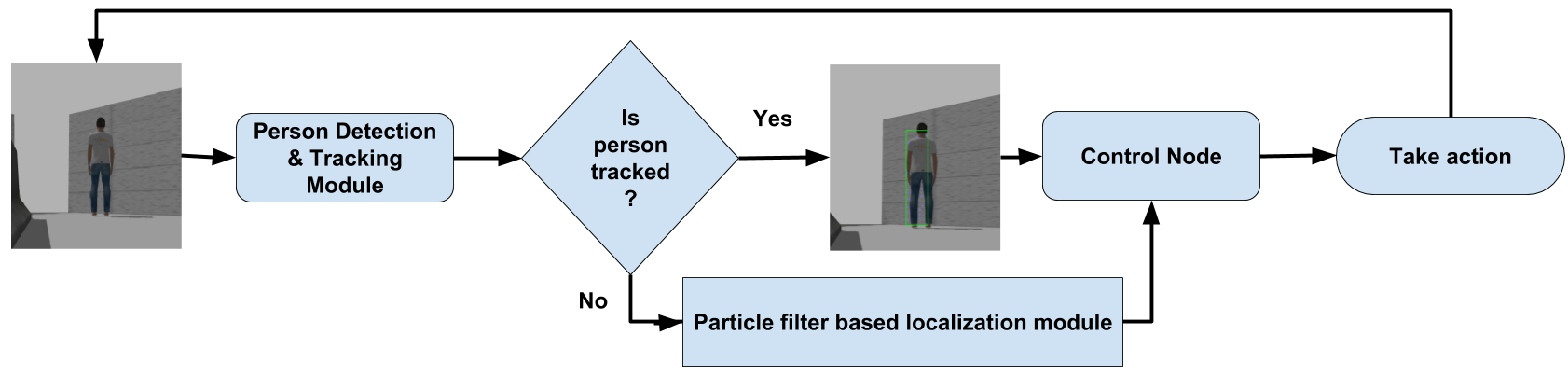}
    \caption{Overview of the pipeline the smart pursuer uses while tracking an evader.}
    \label{fig:2}
\end{figure*}

Initially, we set up the game with a random pursuer that tries to actively track an evader which is also randomly moving in the environment, as a baseline method. This variant of the evader has no knowledge of the actions and observations of the pursuer. Next, we set up the game with our smart pursuer(Fig. ~\ref{fig:2}) to actively track the random evader mentioned above. Lastly, the smart pursuer has to track a smart evader which is aware of the location of the pursuer. We quantify the tracking performances across the scenarios with different pursuer to evader speed ratios in different environments.

\subsection{Object tracking from static camera}
Several methods exist to track an object using a static camera. In our probabilistic game, the pursuer needs to detect the evader and then track it across consecutive frames. As we model the evader agent as a human, we adopt the OpenPose\cite{cao_openpose:_2018} tracker to detect the human. After obtaining a bounding box from the person detection module, we apply the Lucas-Kanade method \cite{bouguet_pyramidal_2000} to track the given image template. Since the pursuer uses a depth camera, it can estimate the depth of the points that relate to the bounding box. The obtained depth of point clouds can be used to approximately determine the location of the evader on the map. There are interesting cases where the evader detection module will fail to provide a bounding box, like when the evader turns at a corner. To address this the pursuer uses particle filters when the detection module fails to estimate the location of the evader on the map. The smart pursuer uses particle filter based localization module to estimate evader location on the map as $p(x_k^e|z_k^p)$.

\subsection{Particle Filter based Tracking}
For tracking effectively, the pursuer needs the state estimate $x^e_k$ of the evader that comprises the location and orientation of the evader on the map. The probability distribution over the state of the evader at time $t$ given the observations from time $\tau$= 0 up to time $t$ is given by $p(x_k|z_1,z_2,z_3..z_k)$. Using Baye's rule we can compute the above value as:
\begin{equation}
\begin{split}
  p(x_k|z_1,z_2 ..z_k) =  \frac{p(x_k|z_1,z_2 ..z_{k-1}) p(z_k|x_k,z_1,z_2 ..z_{k-1})}
{p(z_k|z_1,z_2 ..z_{k-1})} .  
\end{split}
\end{equation}
The measurement $z$ at time $t$ is independent of all previous measurements given the current state and hence the observation model can be simplified to $p(z_k|x_k)$. The above equation can be simplified as follows:
\begin{equation}
    p(x_k|z_1,z_2 ..z_k)= \alpha p(z_k|x_k) \\ p(x_k|z_1,z_2 ..z_{k-1}) ,
\end{equation}
where $p(x_k|z_1,z_2 ..z_{k-1})$ is called the actuation or motion model, and $p(z_k|x_k)$ is the measurement model.

The position of the evader is thus estimated by using the weighted mean of the particles.



\subsection{Smart Evader}

  We propose a smart evader that has information about the pursuer's location on the map and also its orientation. Using this information, the smart evader can compute its goal location in an adversarial manner that can minimize the probability of the pursuer finding it. 

\begin{figure}[h]
    \centering
    \includegraphics[width=0.35\textwidth]{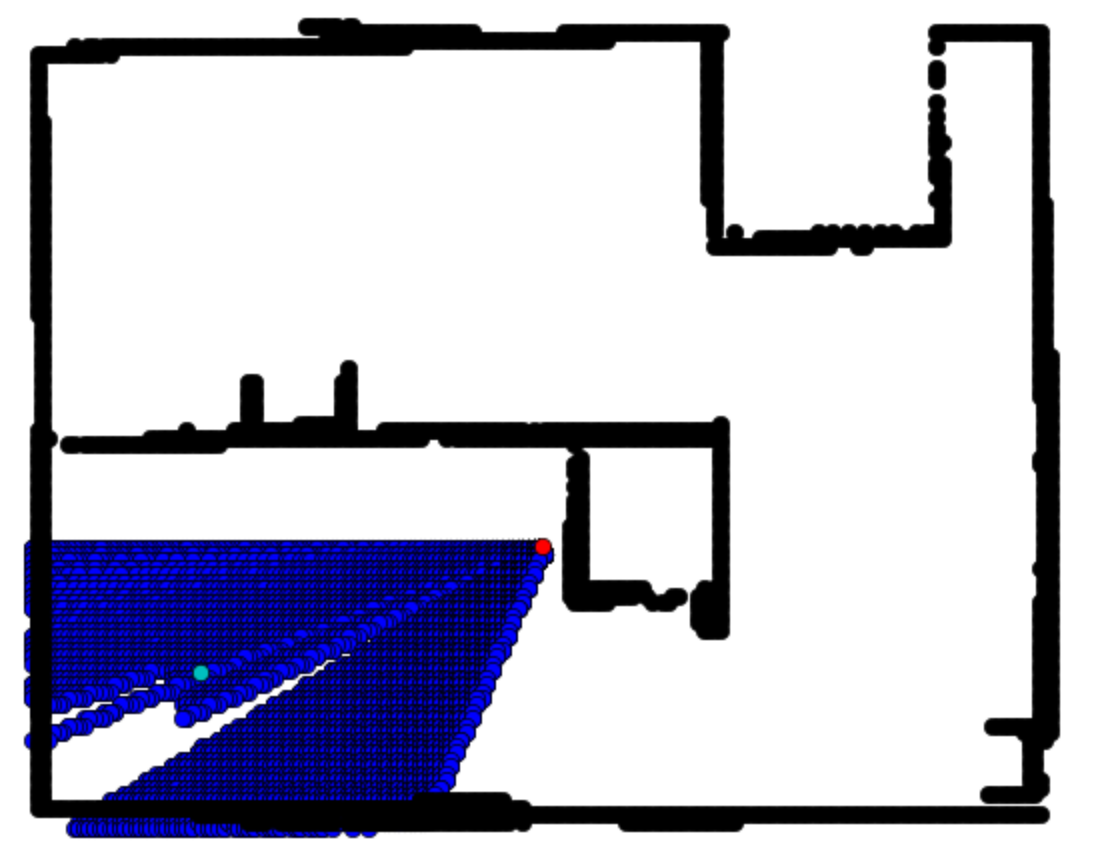}
    \caption{Given position of pursuer (in red) and the orientation, the smart evader can compute visibility region using the ray cast algorithm. The blue dots show the visibility region of the pursuer and the green dot represents position of the evader.}
    \label{fig:3}
\end{figure}

Given the position of the pursuer and it's orientation, the evader can compute the visibility region of the pursuer at time step $k$, $V^p_k$ by using the ray tracing method described in Algorithm \ref{alg:1}. Since the space is discrete, we assume the angle increments: $angle_{step} = 0.0016$ and $distance_{step} = 0.05$ . Let us assume the orientation of the pursuer is $\theta_k^p$ at time step k. Since we know the FOV of the camera used by the pursuer, the rays with orientation $\theta_k^p + \theta_{min, camera}$ and $\theta_k^p + \theta_{max, camera}$ form the boundaries of the visibility region of the pursuer. 

\begin{algorithm}
 \caption{Algorithm to compute visibility region}
 \begin{algorithmic}[1]
 \renewcommand{\algorithmicrequire}{\textbf{Input:}}
 \renewcommand{\algorithmicensure}{\textbf{Output:}}
 \REQUIRE $\{(x_k^p, y_k^p), \theta_k^p, \theta_{min, camera}, \theta_{max, camera} \}$
 \ENSURE  $V^p_k$
 \\ \textit{Initialization} : angle = $\theta_k^p + \theta_{min, camera}$ and $V^p_k = \phi$ 

  \WHILE{angle $ < \theta_k^p + \theta_{min, camera}$}
  \STATE $ dist = dist_{min} $
  \WHILE{ $dist <= dist_{max}$ }
    \STATE $pt = (x + dist * cos(angle), y + dist * sin(angle))$ 
    \STATE add this point to $V^p_k$
    \STATE $dist += distance_{step}$
  \ENDWHILE
    \STATE $angle += angle_{step}$
  \ENDWHILE
  
 \RETURN $V^p_k$
 \end{algorithmic} 
 \label{alg:1}
 \end{algorithm}

After computing the visibility region of the pursuer as shown in Fig. \ref{fig:3}, the evader can now choose a neighbor that maximizes its probability of escape. Let $x^e_{nbr, k}$ be the neighboring locations of the evader and $R^e_k$ be the set of coordinates that are one step away. Here, $R^e_k \subseteq x^e_{nbr, k}$ because some neighbors may contain obstacles and are hence not reachable while some neighbors may be on the other side of an obstacle like a wall. We define a cost function that can compute the importance of a neighbor based on its distance from the pursuer and the evader.

The smart evader, finds the points in free space(coordinates on map without obstacles) $\mathcal{F}$ that are not in visibility region $V^p_k$ of the pursuer as possible locations to escape from the evader. This means the possible goal locations $goal_k^e$ for the smart evader is in $\mathcal{F} \backslash V^p_k$

To iteratively compute the best location for the evader, we consider the distance from the evader to the point in $\mathcal{F} \backslash V^p_k$ and also the distance of the point from the pursuer itself. For the purpose of selecting the best escape location for the evader, we propose a cost function that when minimized will have a better chance of escape for the evader.

Let $cost_{effort, i, k}(goal_k^e[i], (x_k^e, y_k^e))$ be the distance from the evader to a point \textit{i} of the possible locations, $cost_{dist, i, k}(goal_k^e[i], (x_k^p, y_k^p))$ be the distance from the pursuer to a point \textit{i} of the goal points. 

The associated cost($cost_{escape}$) for each point in $goal_k^e$ can be computed using Algorithm~\ref{alg:2}:

\begin{algorithm}
 \caption{Algorithm to compute escape point}
 \begin{algorithmic}[1]
 \renewcommand{\algorithmicrequire}{\textbf{Input:}}
 \renewcommand{\algorithmicensure}{\textbf{Output:}}
 \REQUIRE $\{(x_k^p, y_k^p), (x_k^e, y_k^e), V_{b,k}^p, \mathcal{F} \}$
 \ENSURE  goal
 \\ \textit{Initialization} : $ goal= \phi, cost_{effort} = \phi, cost_{dist} = \phi$
 \FOR{$i= 0$ to length($\mathcal{F} \backslash V^p_k$)}
    \STATE $cost_{escape, i, k} = \frac{cost_{effort, i, k}}{cost_{dist, i, k}}$ 
 \ENDFOR
 \STATE $goal = \mathcal{F} \backslash V^p_k[i]$ s.t $i = \min$ index of $cost_{escape}$
 \RETURN goal
 \end{algorithmic} 
 \label{alg:2}
 \end{algorithm}


\subsection{Smart Pursuer}
Since the random pursuer fails to actively follow the evader, we employ a hybrid approach in the smart pursuer. When the pursuer has visibility of the evader, the pursuer uses a reactive deterministic strategy to keep the evader in the center of the image at every instant of time. The local reactive planner works as follows:
\begin{equation}
    target_{offset\_x} = x_b + w_b / 2 - w_I / 2 .
\end{equation}
Given $(x_b, y_b, w_b, h_b)$ of the bounding box in image and $w_I$ is the width of the image, we compute the offset of the bounding box from the center of the image to help in determining the angular velocity component needed by the pursuer to keep the evader in the center of its line of sight. For the linear component of velocity, we compute the depth of the points that are enclosed in the bounding box region from the depth image and take an average of the total depth. We can compute the linear velocity required to keep at a certain distance from the evader.

In the case when the evader is not visible, we use the particle filter localization module, to estimate the location of the evader and then navigate to that location. We use Sequential Importance Sampling (SIS) algorithm \cite{murata_monte_2015} to implement the particle filter based state estimation of the evader on the map. For the motion model of the evader, we consider a unicycle model and assume the evader cannot translate sideways unless the evader rotates first. The measured forward velocity and robot orientation are represented by variables $V_x$ and $\theta$. The desired speed and direction are represented with the unicycle model using variables V and $\phi$. We can represent the velocity of the unicycle robot as follows:
\begin{equation}
    u = 
    \begin{bmatrix}
        x \\
        y \\
        \phi \\
    \end{bmatrix}
    =
    \begin{bmatrix}
    V * \cos{\phi} \\
    V * \sin{\phi} \\
    \omega 
    \end{bmatrix} .
\end{equation}

For computing the observation likelihood, we reweigh the particles based on the observation of the evader. If the evader is detected by the pursuer, during the weight update stage, the points in the visibility region of the pursuer $V_k^p$ will have more weights when compared to points outside the visibility region since we reduce the weight of points outside visibility region. The opposite happens in the case where the evader is not tracked by the pursuer.

We acknowledge that SIS introduces the degeneracy problem that makes the particles to converge to a single estimate since that particle might have more significant weight than the other particles. Hence we use the effective size $N_{eff}$ of the particle set to mitigate this issue:
\begin{equation}
\begin{split}
    N_{eff} = \frac{1}{\sum_{i=1}^N (w_i)^2}. 
\end{split}
\end{equation}

In the above equation, $N$ is the number of particles in the algorithm and $w_{i}$ is the weight of the particle $i$. If $N_{eff} < \rho . N $ and $ 0 \leq \rho \leq 1 $, then we perform re-sampling to avoid this degeneracy issue. In our experiments we set $\rho = 0.8$



\section{Experiments}

Our proposed framework suggests the following hypotheses that need experimental validation:
\begin{enumerate}
  \item The smart pursuer can track the evader better than a random pursuer.
  \item The smart pursuer maintains a high tracking rate even with comparatively slower speed than the evader.
  \item The smart pursuer maintains a reasonably high tracking performance even when the evader is aware of the intentions of the pursuer.
\end{enumerate}

\begin{figure*}[!htb]
    \centering
    \begin{subfigure}[b]{0.3\linewidth}        
        \centering
        \includegraphics[width=\linewidth]{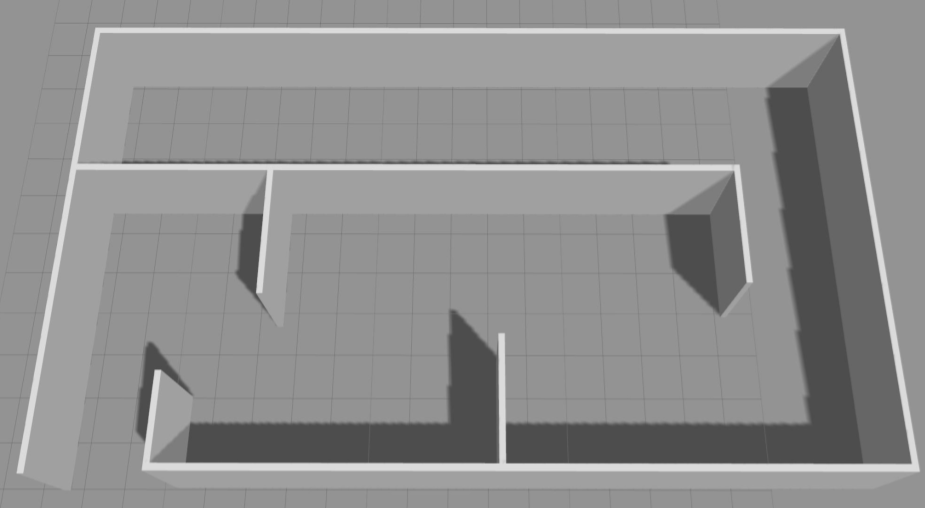}
        \caption{Complex Hall (with no obstacles)}
        \label{fig:4a}
    \end{subfigure}
    \begin{subfigure}[b]{0.3\linewidth}        
        \centering
        \includegraphics[width=\linewidth]{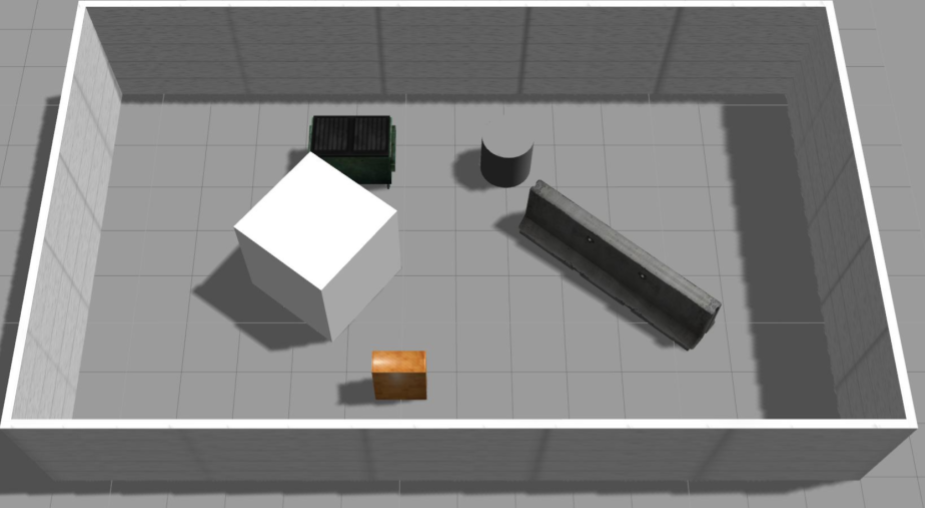}
        \caption{Enclosed Room (with obstacles)}
        \label{fig:4b}
    \end{subfigure}
    \begin{subfigure}[b]{0.3\linewidth}        
        \centering
        \includegraphics[width=\linewidth]{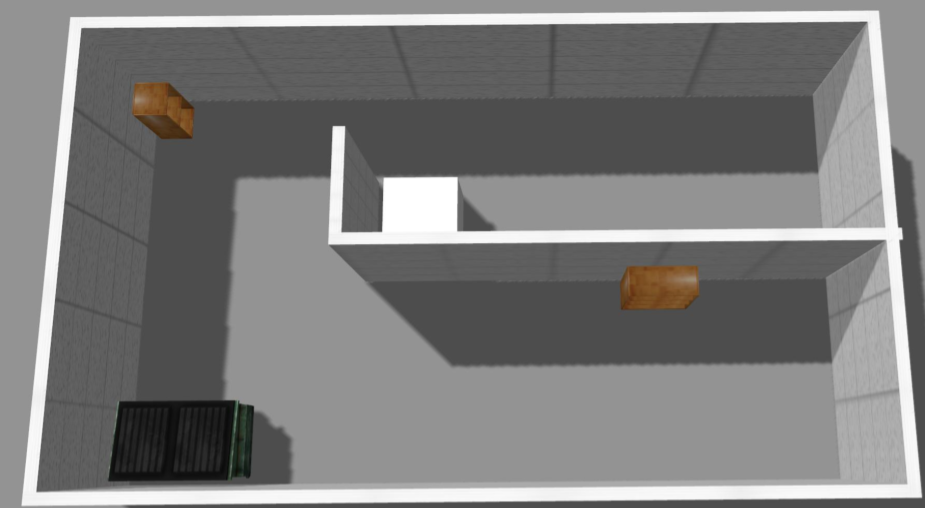}
        \caption{Brick Room (with hall and obstacles)}
        \label{fig:4c}
    \end{subfigure}
    \caption{The three environments used during simulation in Gazebo. The environments vary in complexity and obstacle density.}
    \label{fig:4}
\end{figure*}


  

\subsection{Simulation based validation}


To this end, we evaluate the efficacy of the hypotheses methods in a 3-D indoor environment using Gazebo simulator. We developed the code using ROS and Python, and we make use of \textit{move\_base} package for path planning given a goal location to either the pursuer or evader. The game starts with pursuer detecting the evader and each player take turns to progress in the game. First the evader moves and then pursuer moves. The game stops after $t_{max} = 90 sec$. We implemented the pursuer and evader nodes with a rate of 1Hz so that they have enough time to perform online processing. Since our main concern is the analysis of active object tracking, we provide a comparison of the average success rate of tracking for each scenario.

Previous works on pursuit-evasion games usually utilize a simulated environment where the pursuer and evader are represented using similar robots. For practical implementations, it follows the same method where the evader robot is controlled by a human subject observing a monitor. In our simulations, we use a human model to represent the evader and TurtleBot as the pursuer. For real-world experiments, we leverage Virtual Reality(VR) technology to enable us to overlay information on to the view of the real world. This allows us to model an information game where the human subject can directly play in the game. Our goal in doing so was to bridge the gap between theory and practical implementations by using Hololens to provide data in real-time to the evader to avoid the pursuer better. The human-robot model uses a RGB-D sensor tilted at an angle of $\pi/8$ downward and the parameters like the Field Of View (FOV), near clip and far clip ranges are similar to that of the 3D-sensor on TurtleBot.

To validate the robustness of our approach we created three unique environments that mimic indoor environments of varying complexity. These environments(Fig. ~\ref{fig:4}) differ in the number of obstacles, the density of the obstacles and the layout of the room. The environment with Complex halls(Fig. ~\ref{fig:4a}) is the easiest environment to navigate through since there are no obstacles present. The enclosed environment has multiple obstacles to impede movement, while the brick environment(Fig. ~\ref{fig:4c}) has a wall that allows the evader to hide from the pursuer more effectively. In addition to verifying robustness across multiple environments, we also investigate the effects of various pursuer-to-evader speed ratios.

\subsection{Discussion and Analysis of simulation results}
For each configuration, we run the simulation for 40 iterations, where we spawn the evader and pursuer at various random starting positions in the environment. We present the results for each scenario in our experiment below.

\begin{table*}[!hbt]
  \centering
   \renewcommand{\arraystretch}{1.5}
  \begin{tabular}{|M{2cm}|c|c|c||c|c|c||c|c|c|}
    \hline
    \vfill{\textbf{Pursuer-Evader Speed Ratio ($v_p/v_e$)}} 
    & \multicolumn{3}{c|}{\textbf{Complex Hall}}
    & \multicolumn{3}{c|}{\textbf{Enclosed Room}}
    & \multicolumn{3}{c|}{\centering \textbf{Brick Room}} 
    \\
    \cline{2-10}
    & \textbf{R-R} & \textbf{S-R} & \textbf{S-S} 
    & \textbf{R-R} & \textbf{S-R} & \textbf{S-S} 
    & \textbf{R-R} & \textbf{S-R} & \textbf{S-S}
    \\
    \hline
    $v_p/v_e = 0.5$   & 21.84\% & 55.14\% & 62.33\% & 14.79\% & 55.12\% & 56.67\% & 34.26\% & 51.41\% & 47.42\% \\ \hline
    $v_p/v_e = 1$     & 23.85\% & 78.47\% & 57.6\% & 23.57\% & 60.43\% & 51.30\% & 26.20\% & 59.84\% & 54.31\% \\ \hline
    $v_p/v_e = 2$     & 17.56\% & 84.78\% & 52.65\% & 19.47\% & 78.85\% & 48.08\% & 25.20\% & 69.82\% & 49.52\%  \\ \hline
  \end{tabular}
  \caption{Mean tracking success rate across various environments and speed ratios. R-R represents random pursuer and random evader, S-R represents smart pursuer and random evader, and S-S represents smart pursuer and smart evader.}
  \label{table:1}
\end{table*}

\begin{figure}[!hbt]
    \centering
    \includegraphics[width=\linewidth]{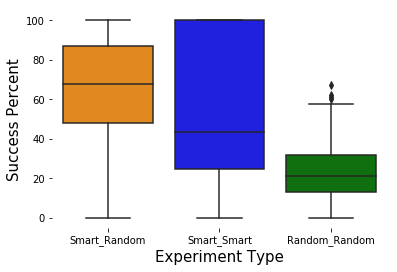}
    \caption{Comparison of overall success rate distributions for different behaviors of the pursuer and evader.}
    \label{fig:8}
\end{figure}

We use boxplots to visualize our data distributions as they can succinctly convey the efficacy of our approach. The box in the boxplot contains $50\%$ of the data; the horizontal line inside the box represents the median of the data, while the top and bottom of the box represent the $75^{th}$ and $25^{th}$ percentile respectively. The vertical lines outside the box show the extent of the data beyond the center of the distribution. Finally, the isolated circles outside the ends of the vertical lines represent the outliers. Fig. ~\ref{fig:6} shows a comparison of the trajectories taken by both the pursuer and evader during simulation. The red path indicates the trajectory of the pursuer and the green path indicates the trajectory of the evader. Fig. ~\ref{fig:8} compares the distribution of success rates for various behaviors of the pursuer and evader. In the case of smart pursuer and smart evader, the boxplot has higher variance of success rate because of the uncertainty of tracking a smart evader that actively tries to escape the pursuer's visibility region.

\subsubsection{Random Pursuer and Evader}
When both the pursuer and evader are random, neither the pursuer nor the evader is actively trying to track or evade its opponent. From the results in Table-1, we can observe that the tracking rate is the worst in this scenario. This makes sense intuitively as since the random pursuer is not actively trying to track an evader, it wanders on the map with the hope of detecting an evader in the future. For random pursuer and random evader trajectory profile, (Fig. ~\ref{fig:6a}), we can observe that both pursuer and evader exhibit random movement and don't try to actively track or evade their opponent. 

\subsubsection{Smart Pursuer and Random Evader}
From Table-1 and Fig. ~\ref{fig:5} we can observe that the tracking performance of a smart pursuer when compared to a random pursuer is significantly higher. This is due to the fact that a smart pursuer actively tries to maximizes its probability of tracking an evader. The trajectory profile of a smart pursuer (Fig. ~\ref{fig:6b}, Fig. ~\ref{fig:6c}) is much less random as the pursuer now tries to actively track the evader. This observation empirically validates our hypothesis (1). 

\subsubsection{Smart Pursuer and Smart Evader}
\begin{figure*}[!htb]
    \centering
    \begin{subfigure}[b]{\linewidth}        
        \centering
       \includegraphics[width=\linewidth]{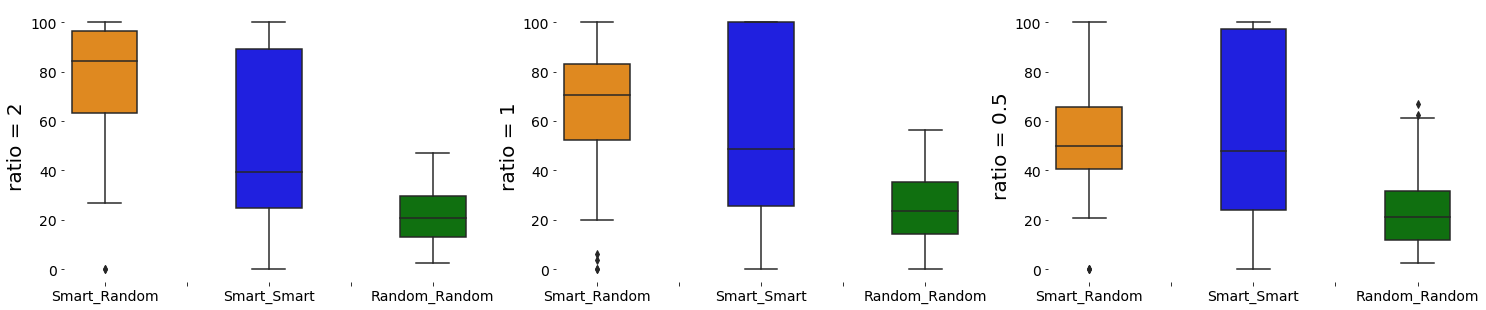}
    \caption{Comparison of the distributions of success rate across different pursuer-to-evader speed ratios. }
    \label{fig:5a}
    \end{subfigure}
    
    \begin{subfigure}[b]{\linewidth}        
        \centering
        \includegraphics[width=\linewidth]{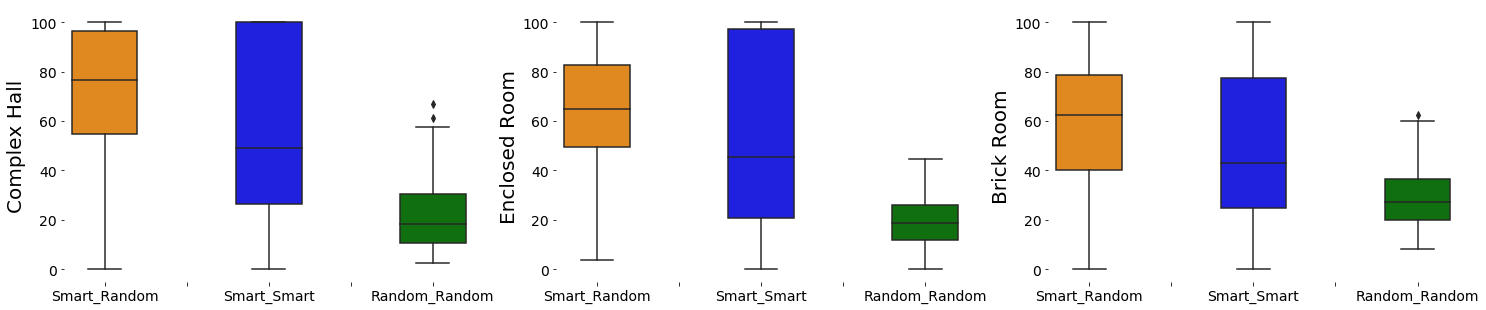}
    \caption{Comparison of the distributions of success rate across different environments.}
    \label{fig:5b}
    \end{subfigure}
    
    \caption{Comparison of the success rate distributions across the various pursuer-to-various speed ratios and environments.}
    \label{fig:5}
\end{figure*}

\begin{figure*}[!t]
    \centering
    \begin{subfigure}[b]{0.3\linewidth}        
        \centering
        \includegraphics[width=\linewidth]{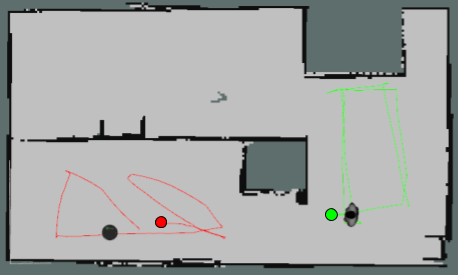}
        \caption{Trajectory for random pursuer and evader}
        \label{fig:6a}
    \end{subfigure}
    \begin{subfigure}[b]{0.3\linewidth}        
        \centering
        \includegraphics[width=\linewidth]{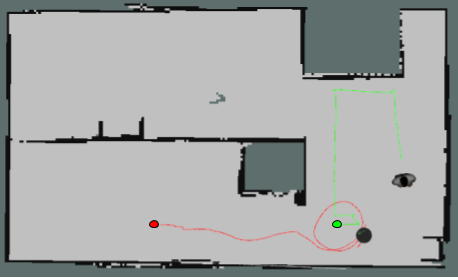}
        \caption{Trajectory for smart pursuer and random evader}
        \label{fig:6b}
    \end{subfigure}
    \begin{subfigure}[b]{0.3\linewidth}        
        \centering
        \includegraphics[width=\linewidth]{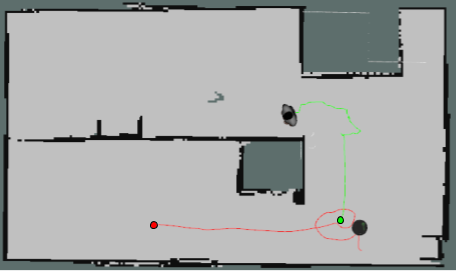}
        \caption{Trajectory for smart pursuer and smart evader}
        \label{fig:6c}
    \end{subfigure}
    \caption{Comparison of the trajectory profiles. The red dot and green dot represent the starting positions of pursuer and evader respectively. As an agent becomes smart, the trajectories become less random and more direct.}
    \label{fig:6}
\end{figure*}

When the evader is smart, it actively tries to escape from the visibility region of the pursuer. From the results in Table-1, we can observe that tracking a smart evader yields lower success rate for the smart pursuer. Also, from Fig. ~\ref{fig:5} we can see that there is increased variance in the tracking performance as compared to the distribution of the smart pursuer and random evader. This can be attributed to the increased uncertainty in the movement of the evader after it is aware of the actions of the pursuer. From the trajectory profile in Fig. ~\ref{fig:6c}, we can see that the smart evader actively tries to move away from the pursuer's visibility region. As aforementioned, compared to the tracking success with random evader, our smart pursuer still maintains a reasonably high success rate, which empirically validates our hypothesis (3).

In addition, from Fig.~\ref{fig:5a}, even with half the speed of both random and smart evader, our smart pursuer still maintains a high tracking success rate, which empirically validates the hypothesis (2).

\subsection{Real experiment based validation}

\begin{figure}[!h]
    \centering
    \includegraphics[width=\linewidth]{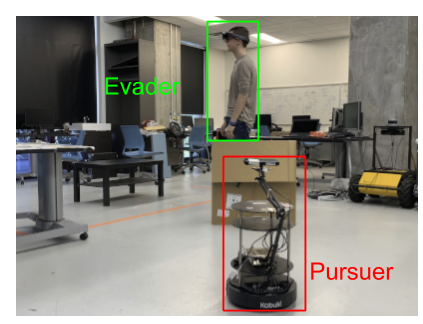}
    \caption{A smart evader accessing adversarial information from Hololens while the pursuer is actively tracking the evader.}
    \label{fig:7}
\end{figure}

For the real experiment, we use TurtleBot as a pursuer along with a human subject acting as an evader. For the test environment, we used our lab which resembles a cluttered workspace with obstacles as shown in Fig.~\ref{fig:7}. We used the TurtleBot to navigate in the environment and created a map of the lab. To mimic the smart evader behavior, we use Hololens to relay information about the map and also the location of the pursuer to the evader so that the evader can avoid the pursuer better. We present the results of this experiment in the video submission.

\section{Related Work}

To further discuss the observed performance from our experiments, we summarize a variety of thrusts of related work in this section. Pursuit-Evasion games can be typically classified into two distinct types: discrete and continuous. This classification is done based on the type of environment the game provides to its agents.

A discrete formulation of this problem is to model the environment as a graph. The players are usually placed on the vertices of a finite graph and the undirected edges that connect the vertices usually denote the possible moves for the player from one vertex to the other.
Parsons \cite{parsons_pursuit-evasion_1978} modeled this problem as a search for a man who is wandering unpredictably on the edges of a finite connected graph. Parsons also introduced the term search number which is equivalent to the minimum number of searchers required to find the lost man.


Megiddo \cite{megiddo_complexity_1988} proved that the complexity of searching a graph, which is equivalent to the search number, is NP-hard. In both these cases, visibility is not considered. The evader is found or captured if and only if both pursuer and evader are on the same vertices of the graph. The cop number is defined as a theoretical limit on the number of pursuers that can be used on a particular graph to successfully capture an evader. The cop number varies for different graphs based on the structure (topology) of the graph. Since we use a single pursuer in our work, we need not consider the cop number, instead we focus on finding a goal location for the pursuer that will maximize it's tracking rate.


In versions of the game where the pursuer is granted additional visibility past its current location, an evader is considered caught if it is present in the field of view of the pursuer. In a polygonal free space, the search number of $\infty$-searchers required is NP-hard\cite{guibas_visibility-based_1999}. Similarly for visibility based version, using a $\phi$-searcher\cite{gerkey_visibility-based_2006} is also NP-hard. Since our game is played on a 2-D grid map and we would not need to convert the occupancy grid into a graph and then play an edge or vertex traversing game. This slightly increases the complexity of the state space of the pursuer.

Using game theory, Hespanha et al. \cite{hespanha_probabilistic_2000} formulated the pursuit-evasion problem as a partial information Markov game, and proposed a Nash solution to the resulting one-step non-zero sum game, provided the evader has access to the pursuer's information. For the visibility based variant, where both pursuer and evader are holonomic with bounded speeds and both have complete knowledge of the map, Bhattacharya et al. \cite{bhattacharya_existence_2009} presented strategies for players that are in Nash equilibrium.


Some researchers have utilized the information about the game environment to create partition regions where the pursuers and evaders might traverse to. One such method \cite{kolling_graph-clear_2007} converts an occupancy grid map to a graph with restrictions by partitioning the workspace. They decompose the occupancy grid using Generalized Voronoi Diagrams to obtain a reduced graph representation of the environment with reduced action space which makes it easy for the pursuer to traverse in the environment.

A Pursuit-evasion game can also be modeled with a probabilistic framework\cite{vidal_probabilistic_2002}. In such a framework, each agent is governed by its unique transition probability function that is dependent on a couple of factors: the agent's actions and an observation probability function that estimates the location of obstacles and the other agent's location. The pursuer agent tries to maximize the probability of capturing the evader at every instant, and the evader tries to minimize the probability of getting captured. Probabilistic pursuit-evasion games may not produce optimal policy solutions that reduce expected capture time, but they can compute an efficient sub-optimal policy with good performance using greedy algorithms. Our work falls into this category, since we model the pursuer as a probabilistic agent that tries to maximize its success rate of tracking the evader.

\section{Conclusion and Future work}
In this paper, we presented a smart pursuer that can track an evader in a known environment with a relatively higher success rate than a random pursuer. We also presented a smart evader that has information about the pursuer's location and orientation and takes an action to reduce the tracking rate of the pursuer. From the experiments, though we showed that the smart evader can avoid the pursuer better in various environments and for different speed ratios of pursuer to evader, our smart pursuer can still track the evader with a reasonably high success rate. Upon the acceptance of the paper draft, we would like to make our source code and our evaluation suite publicly available for future research. 

It is worth mentioning the caveat that we did not further model an even smarter pursuer to take into consideration the fact that: the evader is also smart. We expect a certain equilibrium could be reached under this setting, and our promising initial results suggest future research avenues to study such an equilibrium. Moreover, past work such as \cite{megiddo_complexity_1988} focuses on identifying the number of pursuers required on a graph to successfully capture an evader. The study of multi-robot behaviors for smart pursuers is out of the scope of this work, but it could be a possible and challenging research avenue to pursue in the future.


\bibliography{reference}

\end{document}